\documentclass[letterpaper, 10 pt, conference]{ieeeconf}  

\IEEEoverridecommandlockouts                              

\usepackage{graphics} 
\usepackage{epsfig} 
\usepackage{amsmath} 
\usepackage{amssymb}  

\usepackage[utf8]{inputenc} 
\usepackage[T1]{fontenc}    

\usepackage[dvipsnames]{xcolor}
\usepackage{url}            
\usepackage{booktabs}       
\usepackage{amsfonts}       
\usepackage{nicefrac}       
\usepackage{microtype}      

\usepackage{caption}
\usepackage{subfigure}
\usepackage{booktabs} 
\usepackage{xcolor}
\usepackage{multirow}
\usepackage{pifont}
\usepackage[noend]{algpseudocode}
\usepackage{setspace}
\usepackage{comment}
\usepackage[ruled]{algorithm2e}
\usepackage{verbatim}

\usepackage{wrapfig}
\makeatletter
\let\NAT@parse\undefined
\makeatother
\usepackage{hyperref}
\hypersetup{
    colorlinks=true,
    linkcolor=orange,
    filecolor=magenta,      
    urlcolor=orange,
    citecolor=orange,
}
\usepackage{pdfx}

\title{\LARGE \bf Learning from Imperfect Demonstrations \\ via Adversarial Confidence Transfer}
\author{Zhangjie Cao$^*$, Zihan Wang$^*$, Dorsa Sadigh%
    \thanks{Emails: {\tt\footnotesize caozj@cs.stanford.edu, wangzih@stanford.edu, dorsa@cs.stanford.edu}}%
    \thanks{Computer Science, Stanford University, CA, USA}%
    \thanks{* means Equal Contribution. Author ordering determined by coin flip over a Google Hangout.}
    }

\begin{document}

\maketitle
\thispagestyle{empty}
\pagestyle{empty}


\begin{abstract}
Existing learning from demonstration algorithms usually assume access to expert demonstrations.
However, this assumption is limiting in many real-world applications since the collected demonstrations may be suboptimal or even consist of failure cases.
We therefore study the problem of learning from imperfect demonstrations by learning a confidence predictor.
Specifically, we rely on demonstrations along with their confidence values from a different \emph{correspondent} environment (source environment) to learn a confidence predictor for the environment we aim to learn a policy in (target environment---\emph{where we only have unlabeled demonstrations}).
We learn a common latent space through adversarial distribution matching of multi-length partial trajectories to enable the transfer of confidence across source and target environments.
The learned confidence reweights the demonstrations to enable learning more from informative demonstrations and discarding the irrelevant ones. 
Our experiments in three simulated environments and a real robot reaching task demonstrate that our approach learns a policy with the highest expected return. We show the videos of our experiments on our \href{https://sites.google.com/view/adversarialconfidencetransfer}{\color{orange}{website}}.
\end{abstract}



\section{Introduction}
Standard imitation learning algorithms assume that collected demonstrations are provided by experts, who can always achieve successful outcomes~\cite{abbeel2004apprenticeship,ziebart2008maximum,ho2016generative}. 
However, this assumption is usually violated in real-world applications. Expert demonstrations require domain expertise and thus are expensive to obtain. 
On the other hand, we often have easy access to a plethora of imperfect demonstrations ranging from expert behavior to noise. This is evident when collecting human demonstrations on a robot arm since controlling manipulators with high degrees of freedom can be challenging for human users~\cite{losey2020controlling}, or a bounded rational demonstrator may have difficulty providing optimal demonstrations~\cite{simon1997models,kwon2020humans}.

Prior work on learning from imperfect demonstrations can be roughly categorized into confidence-based and ranking-based methods: Confidence-based methods learn a confidence over demonstrations to re-weight them~\cite{pmlr-v97-wu19a,zhang2021confidenceaware}, while ranking-based methods recover the reward function~\cite{brown2019extrapolating,brown2020better,chen2020learning}. However, all these works require some level of annotation, e.g., confidence or rankings of the demonstrations. For complex environments, large-scale annotations are required to cover the whole state-action space, which is labor-intensive and expensive to obtain. Furthermore, the annotations can often contain a significant amount of noise when collected through crowdsourcing platforms. 

\begin{figure}[t]
\centering
\includegraphics[width=.45\textwidth]{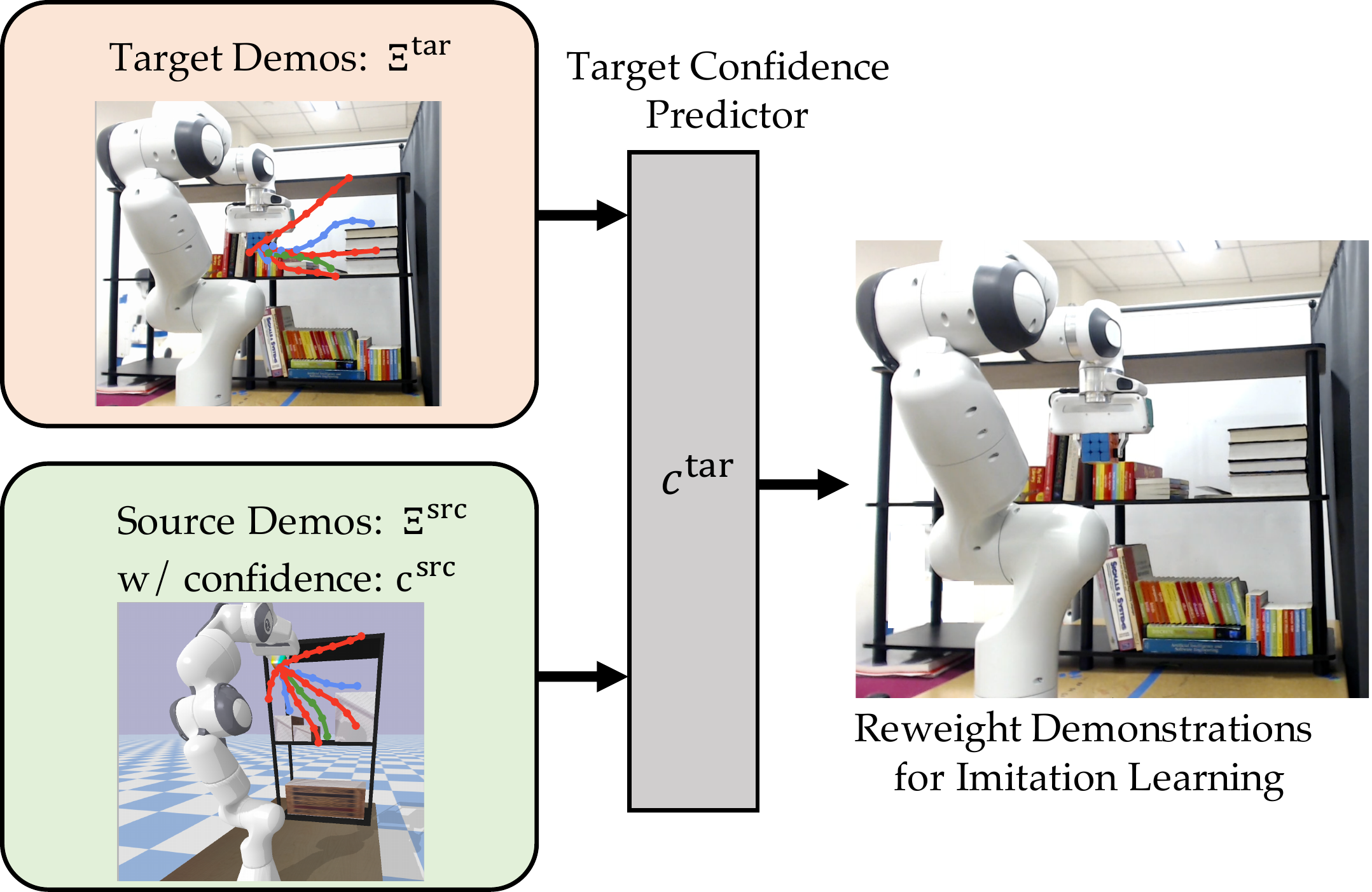}
\caption{\small{Overview of the problem setting.}}
\label{fig:front}
\vspace{-20pt}
\end{figure}

In this paper, we focus on confidence-based methods. As shown in Fig.~\ref{fig:front}, our insight is that instead of using manually annotated confidence over demonstrations, we leverage confidence annotations in a different but related environment---the \emph{source} environment. The source environment may have a different state-action space from the \emph{target} environment---the environment we are operating in. The source environment and the target environment should have a correspondence mapping between state-action pairs~\cite{zhang2021learning}. 
We assume that collecting offline demonstrations along with their confidence annotations is easy in the source environment. 
As an example, the source and target environments can be a simulated robot and a real robot respectively or two different robot arms with different degrees of freedom.
Our key idea is to leverage the correspondence between state-action pairs in the source and target environments to learn a confidence measure without the restrictive assumption of access to confidence annotations of target demonstrations.

This problem setting introduces a new challenge: The source and target environments have different state-action spaces. That means the dimensions of the states and actions and even the semantic meaning of each dimension can be different between the two environments, so we cannot directly apply the source confidence predictor to the target state-action pairs. We instead need to transfer the source confidence predictor to the target environment.

We propose an approach that leverages the assumption of the correspondence between state-action pairs in the source and target environments leading to the existence of a common latent space, where one can learn a shared confidence predictor for both environments.
Our approach aligns both the feature-level distributions and the confidence-level distributions of partial trajectories with different lengths across domains by a domain adversarial loss. We design a multi-length partial trajectory matching, which preserves temporal relationships across consecutive states and actions, and enables an accurate matching of trajectories.

Our contribution is a new confidence-based imitation learning algorithm that learns from imperfect demonstrations. Instead of relying on manual annotations in our environment, we leverage confidence-annotated demonstrations in a related environment---where collecting this data is easier---to learn a confidence predictor for our environment of interest. We experiment with several simulation environments and a real robot reaching task. Our results suggest that our approach achieves both more accurate confidence prediction and higher performance of the imitation policy than other methods.

\section{Related Work}
\label{sec:related_work}

\noindent \textbf{Imitation Learning.} 
Imitation learning aims to learn a policy by imitating the behavior of expert demonstrations, where mainstream methods can be categorized into behavior cloning (BC), inverse reinforcement learning (IRL), and generative adversarial imitation learning (GAIL). BC directly learns the policy mapping state to action by supervised learning over state-action pair data in the demonstrations~\cite{bain1995framework,ross2010efficient,ross2011reduction}. IRL first recovers the reward function from demonstrations and learns a policy using reinforcement learning based on the learned reward function~\cite{ng2000algorithms,abbeel2004apprenticeship,ziebart2008maximum}. GAIL aligns the occupancy measure between the policy and the demonstrations with adversarial learning~\cite{ho2016generative,fu2017learning,xiao2019wasserstein}. These standard imitation learning methods usually assume that all of the demonstrations are from experts who successfully finish the task. This work goes beyond this assumption by allowing for learning from suboptimal or even malicious demonstrations.

\noindent \textbf{Imitation Learning from Imperfect Demonstrations.} 
Imitation learning from imperfect demonstrations aims to recover a well-performing policy from demonstrations that can range from random noise to optimal. Ranking-based and confidence-based methods are the two core approaches to address this problem. Ranking-based methods recover the reward function from provided ground-truth~\cite{brown2019extrapolating,brown2019deep}, self-generated~\cite{brown2020better,chen2020learning}, or interactively collected rankings~\cite{myers2021learning,novoseller2020dueling}. In our setting, we can only generate rankings and learn the reward in the source environment, which cannot be applied to the target environment with a different state space. Target rankings can be self-generated by injecting noise in the target demonstrations as in \cite{brown2020better}, but that would require the self-generated demonstrations to be ranked worse than the original ones. This may not hold in our setting with the potential of nearly random demonstrations in our dataset.  

Confidence-based methods typically learn the confidence predictor from confidence annotations~\cite{pmlr-v97-wu19a,zhang2021confidenceaware} but the annotations are assumed to be on target demonstrations. We go beyond this assumption in our setting, since the confidence predictors are from a different environment. Cao et al. learn the target confidence predictor from the reward of demonstrations in the source environment but still assume the two environments share the same state space~\cite{cao2021learning}. We target a more general setting, where the source and target environments have different state-action spaces and dynamics.

Other than learning from suboptimal demonstrations, prior works also consider learning from demonstrations with different dynamics and develop feasibility metrics that learn from this data~\cite{cao2021learning,cao2021learning1}. However, unlike this paper, these works require the same state space across the two dynamics.

\noindent \textbf{Correspondence of Dynamics.}
To learn the correspondence~\cite{nehaniv2002correspondence} between different dynamics, there are mainly two categories of methods: learning a translation mapping~\cite{taylor2007transfer}, or a common latent space~\cite{sadeghi2016cad2rl, 8794310} for states and actions. To learn a translation mapping, Tzeng et al.~\cite{tzeng2020adapting} weakly align pairs of images with paired image observations. Ammar el al.~\cite{ammar2015unsupervised} utilize unsupervised manifold alignment to find correspondence between states of demonstrations. Kim et al.~\cite{kim2019cross} and Zhang et al.~\cite{zhang2021learning} recently translated states and actions across environments by learning a unidirectional mapping of states and a cycle of forward mapping and inverse mapping between actions. 
Learning a translation mapping is often much more challenging than simply finding a common latent space.
A common latent space is sufficient for the transfer of a confidence predictor across different dynamics.

To learn a common latent space, domain randomization methods align the different dynamics by randomizing the dynamics with hand-crafted augmentations~\cite{tobin2017domain,peng2018sim,zakharov2019deceptionnet}, but they require the differences between the two dynamics to be covered by the augmentations. Gupta et al.~\cite{gupta2017learning} require pairs of states from different dynamics to learn the latent space. The assumptions of paired data or covering augmentations are often restrictive in practice. We remove these assumptions and only assume the existence of correspondence and learn the invariant representation space using multi-length partial trajectory matching.

\section{Problem Statement}
\label{sec:problem}
Let the source and target environments be represented by two deterministic Markov decision processes (MDP), $M^\text{src}$ and $M^\text{tar}$, where $\mathcal{M}^\bullet= \langle \mathcal{S}^\bullet, \mathcal{A}^\bullet, p^\bullet, \mathcal{R}^\bullet, \rho_0^\bullet, \gamma^\bullet \rangle, \bullet\in\{\text{src},\text{tar}\}$. $\bullet=\text{src}$ indicates the source domain, while $\bullet=\text{tar}$ indicates the target domain. $\mathcal{S}^\bullet$ is the state space, $\mathcal{A}^\bullet$ is the action space, $p^\bullet: \mathcal{S}^\bullet \times \mathcal{A}^\bullet  \rightarrow\mathcal{S}^\bullet $ is the transition function, $\rho_0^\bullet$ is the initial state distribution, $\mathcal{R}^\bullet: \mathcal{S}^\bullet \times \mathcal{A}^\bullet \rightarrow \mathbb{R}$ is the reward function, and $\gamma^\bullet$ is the discount factor. A policy for $M^\text{tar}$ is $\pi^\text{tar}: \mathcal{S}^\text{tar} \times \mathcal{A}^\text{tar} \rightarrow [0,1]$ defining a probability distribution over the target action space in a given state. To evaluate a policy $\pi^\text{tar}$, we use the expected return, which is defined as $\eta_{\pi^\text{tar}}= \mathbb{E}_{s_0 \sim \rho_0^\text{tar},\pi^\text{tar}}\left[\sum_{t=0}^{\infty}(\gamma^\text{tar})^{t} \mathcal{R}^\text{tar}(s_{t}, a_{t}, s_{t+1})\right]$, where $t$ indicates the time step.
A trajectory $\xi^\bullet$ is drawn from either the source or target environments, and is a sequence of state-action pairs: $\xi^\bullet=\{s^\bullet_0,a^\bullet_0,s^\bullet_1,\dots,a^\bullet_{T-1},s^\bullet_T\}$.

\noindent \textbf{Problem Definition.}
In our problem setting, we are given a set of imperfect target demonstrations $\Xi^\text{tar}=\{\xi^\text{tar}_1, \xi^\text{tar}_2,\dots\}$ in the target environment $M^\text{tar}$, and a set of imperfect source demonstrations $\Xi^\text{src}=\{\xi^\text{src}_1, \xi^\text{src}_2,\dots\}$ along with a measure of confidence $c^\text{src}$ in the source environment $\mathcal{M}^\text{src}$, where the confidence function $c^\text{src}:\mathcal{S}^\text{src}\times \mathcal{A}^\text{src} \rightarrow \mathbb{R}$ evaluates how useful the input state-action pairs are and if they are available or easy to access. Our goal is to find a well-performing policy $\pi^\text{tar}$ for the target environment. Note that we go beyond standard imitation learning by allowing these demonstrations in both environments to range from useless or even harmful demonstrations to nearly optimal or fully optimal ones. At the high level, our approach is to learn a confidence function $c^\text{tar}:\mathcal{S}^\text{tar}\times \mathcal{A}^\text{tar} \rightarrow \mathbb{R}$, which assigns a value to how confident we are in the optimality of the action $a^\text{tar}_t$ in a particular state $s^\text{tar}_t$. We can then reweight the target state-action pairs and conduct standard imitation learning on the reweighted state-action pairs to learn a policy.
To learn $c^\text{tar}$, we would like to learn a confidence predictor $f^\text{tar}$ for the target environment by leveraging knowledge of the confidence function $c^\text{src}$ from the source environment.

\noindent \textbf{Assumptions.}
To transfer the confidence predictor from the source to the target environment, we assume that the source and target environments are \emph{correspondent}, which means that there exists a relational mapping between the state-action pairs of the source agent and the target agent~\cite{zhang2021learning}. Specifically, there exists a relation on states: $\Phi:\mathcal{S}^\text{src}\times \mathcal{S}^\text{tar}$. There also exists a mapping from source state-action pairs to the target action space, $H_1:\mathcal{S}^\text{src}\times \mathcal{A}^\text{src}\rightarrow \mathcal{A}^\text{tar}$, and another mapping from target state-action pairs to the source action space, $H_2:\mathcal{S}^\text{tar}\times \mathcal{A}^\text{tar}\rightarrow \mathcal{A}^\text{src}$. The mappings $\Phi$, $H_1$, and $H_2$ satisfy the following: 
if $(s^\text{src},s^\text{tar})\in \Phi$, then $\forall a^\text{src}\in \mathcal{A}^\text{src}, (s_\text{next}^\text{src}, s_{H_1}^\text{tar})\in \Phi$ and $\forall a^\text{tar}\in \mathcal{A}^\text{tar}, (s_\text{next}^\text{tar}, s_{H_2}^\text{src})\in \Phi$. Here, $s_\text{next}^\text{src}$ is the successor state from $s^\text{src}$ when taking the action $a^\text{src}$, and $s_{H_1}^\text{tar}$ is the successor state from state $s^\text{tar}$ when taking the action $H_1(s^\text{src},a^\text{src})$, and $s_\text{next}^\text{tar}$ and $s_{H_2}^\text{src}$ are defined similarly.
The assumption builds a connection between the source and target agents. Without this assumption, the two environments may have no relations to each other, and there will not be any reason for knowledge transfer across them. This assumption is often satisfied in many real-world applications. Specifically, we are interested in two common settings: 
(1) The source environment being a simulation of a real robot that would define the target environment.
For example, the source environment can be a simulated robotic arm picking an apple from a bowl in simulation, while the target environment can be a real robotic arm picking an apple from a bowl. (2) The source and target environments are different robots performing the same task in a similar context. For example, the source environment can be a simpler 3-DoF robot arm reaching the center of a bowl without colliding with obstacles, while the target robot can be a 7-DoF robot performing the same task. 
We note that the correspondence assumption is less restrictive compared to assumptions in related prior work such as assuming the same dynamics across the environments~\cite{pmlr-v97-wu19a,brown2020better}. 

\begin{figure*}
    \centering
\includegraphics[width=\textwidth]{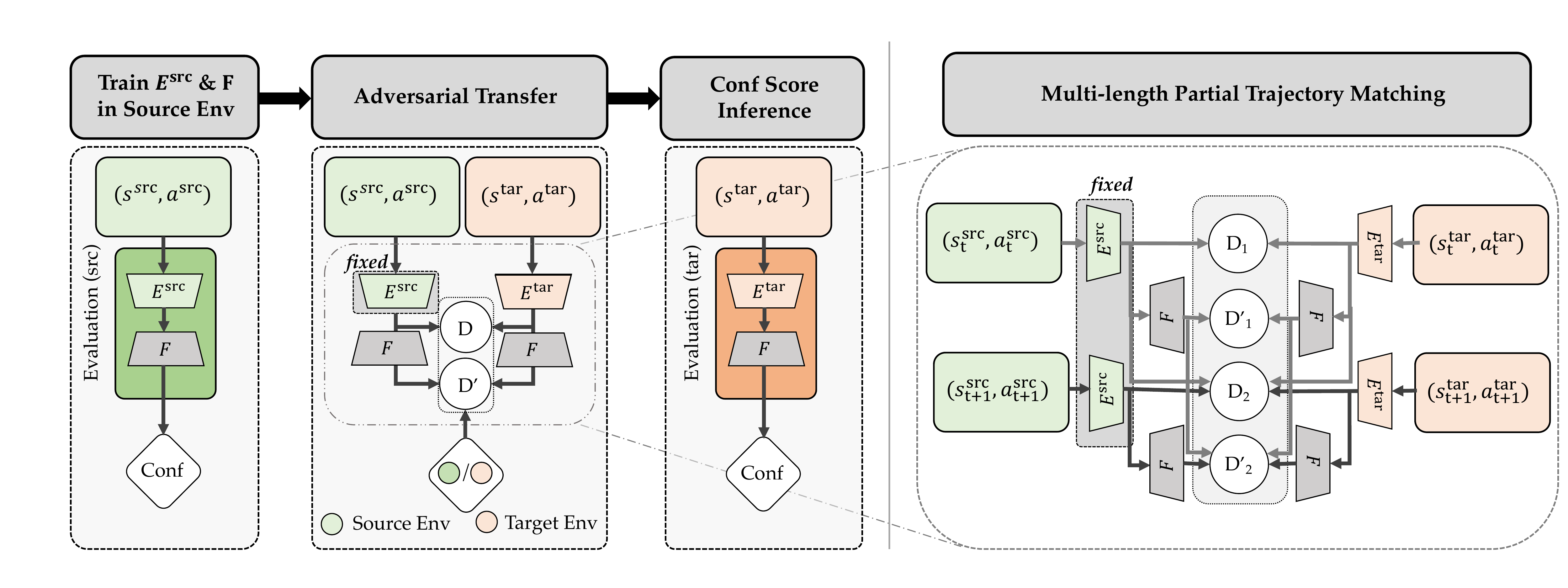}
\vspace{-30pt}
    \caption{\small{(left) The overview of the learning process consisting of three stages: In the first stage, we train the source encoder $E^\text{src}$ and the shared decoder with source confidence-labeled demonstrations $s^\text{src},a^\text{src}$. In the second stage, we align the latent distributions of source and target state-action pairs with a multi-length partial trajectory matching, where we explain the second stage in detail in the right figure. In the third stage, we use the learned $E^\text{tar}$ chained with $F$ as the target confidence predictor. (right) Multi-length partial trajectory matching: we align the feature-level distributions (output of encoders $E^\text{src}$ and $E^\text{tar}$) of different lengths' partial trajectories with different discriminators $D_1,D_2,\dots, D_n$ and align the confidence-level distributions (output of the shared decoder $F$) of different lengths' partial trajectories with different discriminators $D'_1,D'_2,\dots, D'_n$.}}
    \label{fig:overview}
    \vspace{-20pt}
\end{figure*}

\section{Adversarial Confidence Transfer}
\label{sec:method}
We aim to leverage the confidence annotation in the source environment and transfer the confidence knowledge to the target environment. Our key insight is to map the source and target state-action pairs to a common latent space $\mathcal{Z}$, and enforce similarity between the latent features ($z^\text{src}$ and $z^\text{tar}$) for correspondent state-action pairs.
Here, $(s^\text{src}, a^\text{src})$ and $(s^\text{tar}, a^\text{tar})$ satisfy $(s^\text{src}, s^\text{tar})\in \Phi$ and $(s_\text{next}^\text{src}, s_\text{next}^\text{tar}) \in \Phi$.
We then build a shared decoder to predict the confidence from the latent features. 

Since the source and target environments have different state-action spaces, we learn two different encoders $E^\text{src}:\mathcal{S}^\text{src}\times \mathcal{A}^\text{src}\rightarrow \mathcal{Z}$ and $E^\text{tar}:\mathcal{S}^\text{tar}\times \mathcal{A}^\text{tar}\rightarrow \mathcal{Z}$ to map the source and target state-action pairs into a common latent space $\mathcal{Z}$.
We then learn a shared decoder $F:\mathcal{Z}\rightarrow \mathbb{R}$ to predict the confidence from latent features with confidence-labeled demonstrations in the source environment. The target confidence predictor can be computed by $F \circ E^\text{tar}$. Now, the challenge is how to learn the encoders to ensure that the common latent space satisfies the above requirements.

\subsection{Multi-length Partial Trajectory Matching}\label{sec:multi-length}
To map correspondent state-action pairs to a similar latent feature, we develop a distribution matching objective to align the latent distribution of source and target state-action pairs. We can directly align the distribution of state-action latent features, i.e., matching $E^\text{src}(s^\text{src}_t,a^\text{src}_t)$ and $E^\text{tar}(s^\text{tar}_t,a^\text{tar}_t)$. 
However, this does not capture the temporal dependency between state-action pairs over a trajectory.
To address this issue, we incorporate a prior that captures this temporal dependency when learning the encoders. 
Imagine a setting where a simulated robot and a real robot are both making a burger. The state-action pairs corresponding to placing the ingredients (e.g., the patty, lettuce, and tomatoes) are always after placing the bottom bun. If we only match the state-action pairs one at a time between the simulated and the real robot, placing the bottom bun may match placing any of the ingredients or even the top bun, but if we also match the consecutive steps of the state-action pairs, placing the bottom bun in both environments will more likely capture the information that the bottom bun comes before the rest of the ingredients.
So, as shown in Fig.~\ref{fig:overview} (right), we propose a multi-length partial trajectory alignment that emphasizes learning the temporal relationship between state-action pairs.
Specifically, we match the latent feature distribution of length $k$, $(k=1,2,\dots,K)$ partial trajectories, which preserves the temporal relationship between consecutive states and actions, and make the latent features of correspondent state-action pairs more likely to be aligned.

\subsection{Feature-Level and Confidence-Level Matching}\label{sec:matching-loss}
We can now develop our learning algorithm that aligns the latent distribution of state-action pairs. We first train the source encoder $E^\text{src}$ and the shared decoder $F$ with source confidence-labeled state-action pairs using a regression loss:
\begin{equation}\label{eqn:reg}
\begin{small}
\begin{aligned}
    \mathcal{L}_\text{reg} =  \mathbb{E}_{(s^\text{src},a^\text{src})\sim\xi^\text{src},\xi^\text{src} \sim \Xi^\text{src}}\left[\lVert F(E^\text{src}(s^\text{src},a^\text{src})) -c^\text{src}(s^\text{src},a^\text{src})\rVert\right].
\end{aligned}
\end{small}
\end{equation}
After learning $E^\text{src}$ and $F$, we fix the parameters of $E^\text{src}$, and $F$ and use the latent feature space of $E^\text{src}$ as the common feature space. We then only need to make the target encoder $E^\text{tar}$ encode the target state-action pairs into this shared feature space and match the source latent feature distribution. 

We learn the latent feature distribution alignment using a generative adversarial network. Since we only want the partial trajectories of the same length to be matched, for each length of partial trajectories $k$, we adopt a discriminator $D_k$ and a loss $\mathcal{L}^k_\text{fea}$ to match the latent distribution of length-$k$ partial trajectories. 
Specifically, the discriminator $D_k$ aims to discriminate the latent features of source and target length-$k$ partial trajectories, while the target encoder $E^\text{tar}$ is trained to confuse the discriminator. The loss for the discriminator $D_k$ can be derived as a binary classification loss:
\begin{equation}\label{eqn:aligning}
\begin{small}
\begin{aligned}
    \mathcal{L}^k_\text{fea}= 
    &-\mathbb{E}_{(s_1,a_1,\dots,s_k,a_k)\sim \xi, (\xi, \dot) \sim \Xi^\text{src}}\\
    & \left[\log\left(D_k(\text{Cat}\left(E^\text{src}(s_1,a_1), \dots, E^\text{src}(s_k,a_k)\right)\right)\right] \\
    & - \mathbb{E}_{(s_1,a_1,\cdots,s_k,a_k)\sim \xi, \xi \sim \Xi^\text{tar}}\\
    &\left[\log\left(1-D_k(\text{Cat}\left(E^\text{tar}(s_1,a_1), \dots, E^\text{tar}(s_k,a_k)\right)\right)\right].
\end{aligned}
\end{small}
\end{equation}
We concatenate the encoded features of multiple consecutive state-action pairs as the latent feature for the length-$k$ partial trajectories. The discriminator $D_k$ is trained to minimize the discriminator loss $\mathcal{L}^k_\text{fea}$ and the target encoder $E^\text{tar}$ is trained to maximize the discriminator loss. 

However, the above losses only match the latent feature distributions of state-action pairs, which is confidence-agnostic. So there may still exist mismatches of state-action pairs with different confidence values. As shown in Fig.~\ref{fig:overview} (right), to address the issue, we introduce a confidence-level distribution matching to match the predicted confidence of source and target state-action pairs, which further guides the latent features to be matched in a confidence-aware way. Specifically, we use another discriminator $D'_k$ for matching the length-$k$ confidence sequence of length-$k$ partial trajectories using a similar binary classification loss:
\begin{equation*}
\begin{small}
\begin{aligned}
    \mathcal{L}^k_\text{con}=&-\mathbb{E}_{(s_1,a_1,\cdots,s_k,a_k)\sim \xi, (\xi, \dot) \sim \Xi^\text{src}} \\
    &\left[\log\left(D'_k(\text{Cat}\left(F(E^\text{src}(s_1,a_1)), \dots, F(E^\text{src}(s_k,a_k))\right)\right)\right] \\
    & - \mathbb{E}_{(s_1,a_1,\dots,s_k,a_k)\sim \xi, \xi \sim \Xi^\text{tar}} \\
    &\left[\log\left(1-D'_k(\text{Cat}\left(F(E^\text{tar}(s_1,a_1)), \dots, F(E^\text{tar}(s_k,a_k))\right)\right)\right].
\end{aligned}
\end{small}
\end{equation*}
Similarly, $D'_k$ is trained to minimize the discriminator loss $\mathcal{L}^k_\text{con}$ and $E^\text{tar}$ is trained to maximize the discriminator loss. 
\begin{wrapfigure}{r}{.26\textwidth}
\vspace{-10pt}
\begin{equation}\label{eqn:optim_reg}
\begin{small}
\min_{E^\text{src},F}\mathcal{L}_\text{reg}
\end{small}
\end{equation}
\begin{equation}\label{eqn:optim_dis}
\begin{small}
\min_{D_k,D'_k} \mathcal{L}^k_\text{fea}+\mathcal{L}^k_\text{con} (k\in[1,K])
\end{small}
\end{equation}
\begin{equation}\label{eqn:optim_enc}
\begin{small}
\max_{E^\text{tar}} \sum_{k=1}^{K}\left(\mathcal{L}^k_\text{fea}+\lambda \mathcal{L}^k_\text{con}\right)
\end{small}
\end{equation}
\vspace{-18pt}
\end{wrapfigure}To optimize the whole network, we first train the source encoder and the shared decoder with the source confidence-labeled data as shown in Eqn.~\eqref{eqn:optim_reg} (the first stage in Fig.~\ref{fig:overview} (left)). We then iteratively train the feature-level and confidence-level discriminators and the target encoder as shown in Eqn.~\eqref{eqn:optim_dis} and Eqn.~\eqref{eqn:optim_enc}, where $\lambda$ is the trade-off between $\mathcal{L}_\text{fea}$ and $\mathcal{L}_\text{con}$ (the second stage in Fig.~\ref{fig:overview} (left)). After convergence, we chain the target encoder $E^\text{tar}$ and the confidence decoder $F$ to form the target confidence predictor as shown under \textbf{Conf. Score Inference} in Fig.~\ref{fig:overview}.

We can then predict the confidence of each state-action pair $(s,a)$ in the target environment as $F(E^\text{tar}(s,a))$. With the confidence of each state-action pair, when conducting imitation learning on the target demonstrations, we use the confidence to re-weight each state-action pair in the imitation loss as in~\cite{pmlr-v97-wu19a}. Then the imitation learning policy can learn more from useful demonstrations with higher confidence.

  

\noindent \textbf{Relation to Existing Works.}
Our method relaxes assumptions in prior works~\cite{pmlr-v97-wu19a,cao2021learning} and targets a more realistic problem.
However, we argue that our work is compatible with prior works. The common latent feature space in our work can be regarded as a new space of state-action pairs for the source and target environments, where the correspondence between state-action pairs is preserved, especially w.r.t. confidence. With the common latent feature space, we can easily extend prior works to the setting of different state spaces by conducting the algorithm, that is originally performed on states, on the latent feature. Our contribution is not only providing a new approach to learning from imperfect demonstrations but also expanding the usage of algorithms from prior work by introducing a new space to operate in.


\section{Experimental Results}
We conduct experiments on two MuJoCo environments, a simulated and a real Franka Panda arm, and have released the code \href{https://www.dropbox.com/s/8pi4yrfb8bricxn/code.zip?dl=0}{\color{orange}{here}}. We compare our method (\textbf{Ours}) with baselines \textbf{GAIL}~\cite{ho2016generative}, Dynamics Cycle-Consistency (\textbf{DCC})~\cite{zhang2021learning}, and variants of our method by gradually adding modules: starting from GAIL without confidence weighting, we first add feature-level matching as \textbf{Ours-Feature}. We then add the confidence-level matching as \textbf{Ours-Confidence}. Adding multi-length partial trajectory matching to Ours-Confidence derives \textbf{Ours}. 
We show the results of using the ground-truth confidence to reweight the target demonstrations (\textbf{Oracle}). The details about our baseline implementations are at our \href{https://sites.google.com/view/adversarialconfidencetransfer
}{\color{orange}{website}}.

For the Mujoco environments, we evaluate the average return of $100$ rollouts. For both the simulated robot and sim-to-real environments, we evaluate the average return of $500$ rollouts. In all the experiments, we compute the results over $10$ runs and report the mean and the standard deviation. We compute the p-values using the student's t-test. The details on the process of demonstration generation for all environments are shown on our \href{https://sites.google.com/view/adversarialconfidencetransfer
}{\color{orange}{website}}.

\begin{figure}[ht]
\vspace{-10pt}
    \centering
    \subfigure[1-joint]{\includegraphics[width=.115\textwidth]{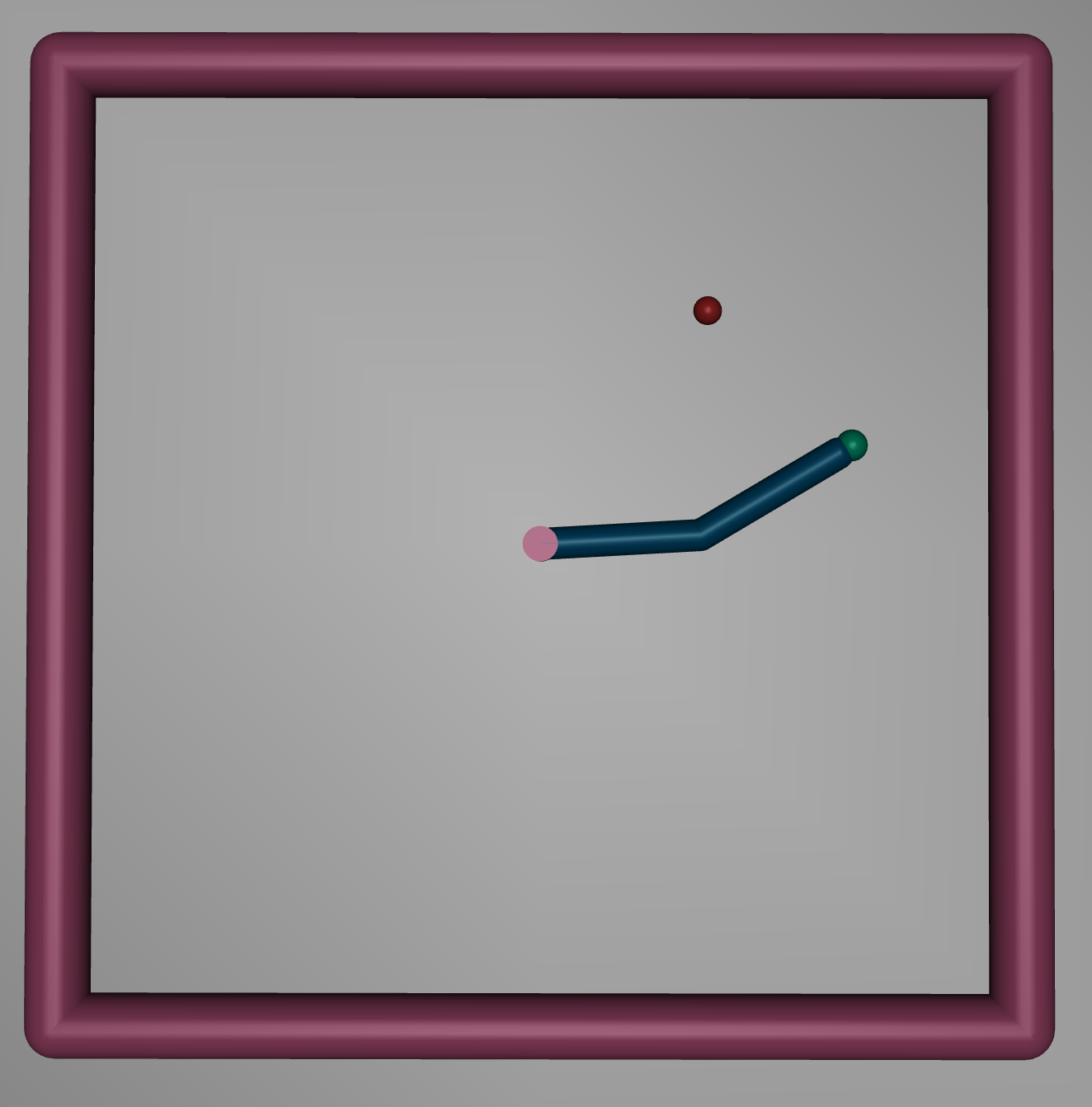}\label{fig:2_joint_reacher}}
    \subfigure[2-joint]{\includegraphics[width=.115\textwidth]{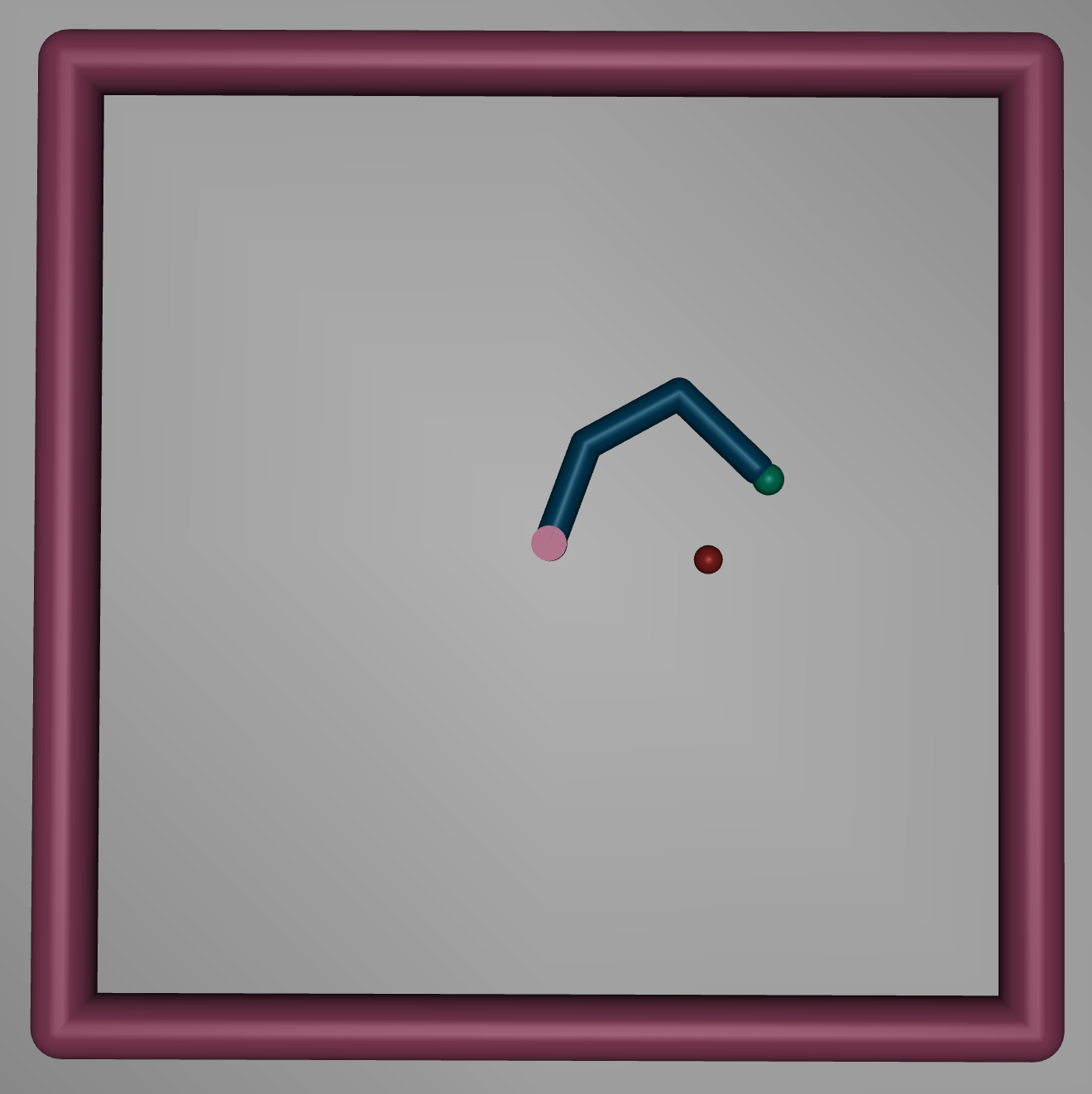}\label{fig:3_joint_reacher}}
    \subfigure[4-leg]{\includegraphics[width=.115\textwidth]{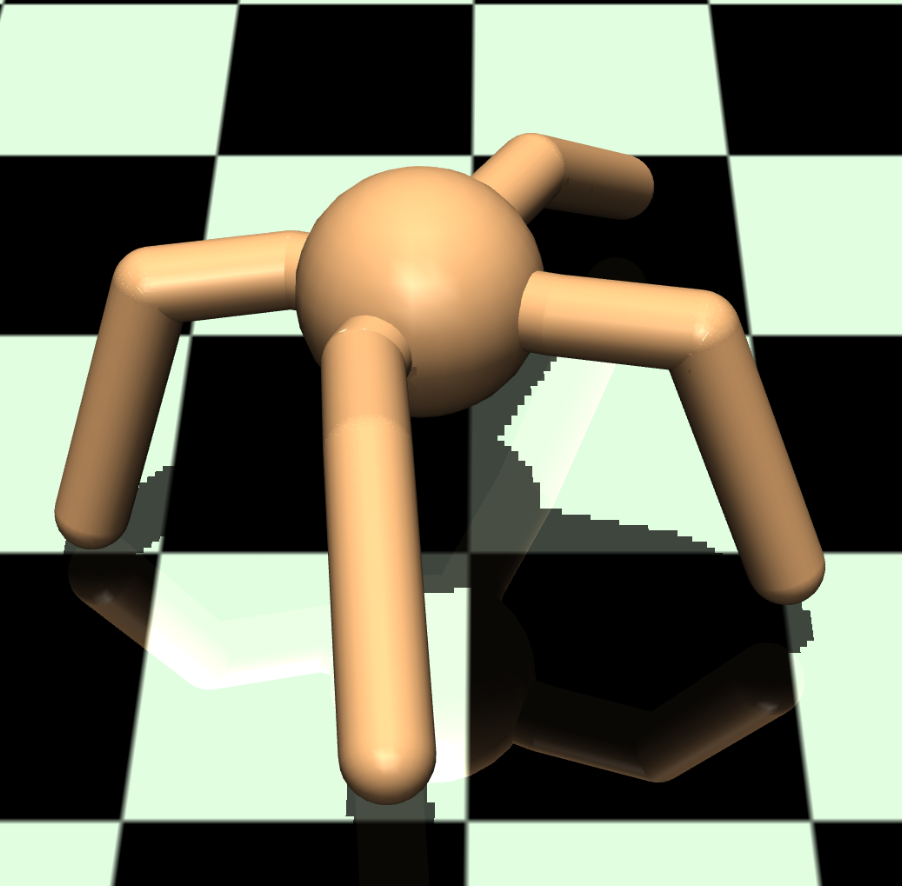}\label{fig:4_leg_ant}}
    \subfigure[5-leg]{\includegraphics[width=.115\textwidth]{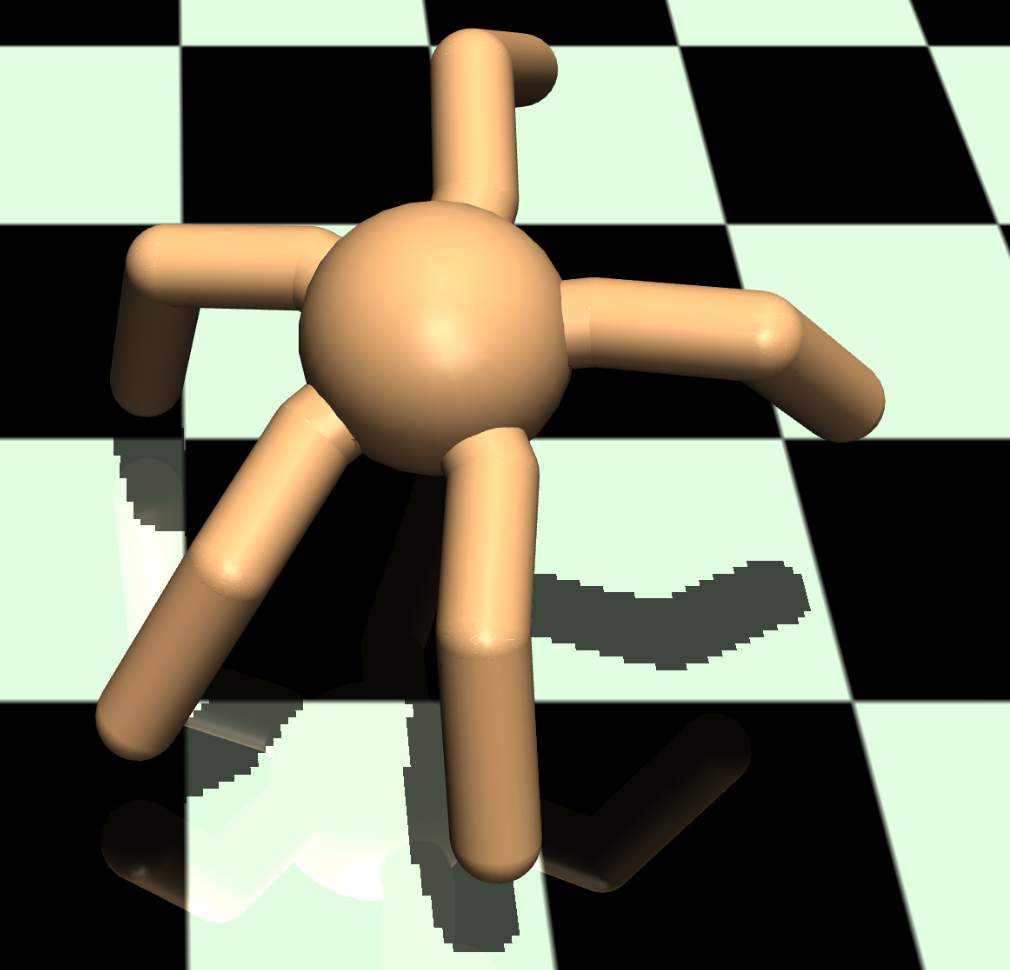}\label{fig:5_leg_ant}}
    \vspace{-10pt}
    \caption{Illustration of source and target MuJoCo environments. The left two figures are Reacher and the right two are Ant.}\label{fig:mujoco}
    \vspace{-10pt}
\end{figure}

\noindent \textbf{MuJoCo Environment.}
We create 4 different MuJoCo environments (see Fig.~\ref{fig:mujoco}): 1-joint and 2-joint reacher, 4-leg and 5-leg ant. 1-joint reacher and 4-leg ant are the source environments, and the 2-joint reacher and 5-leg ant are the target environments. The task for the reacher is to reach the red point from its initial configuration. The task for the ant is to move towards the right horizontally as fast as possible. We train the optimal policy using the RL algorithm TRPO~\cite{schulman2015trust}. For the Reacher environment, we select the random policy, a partially-trained policy, and the optimal policy to collect demonstrations with mixed confidence. For the Ant environment, we select the random policy, three partially-trained policies, and the optimal policy to collect demonstrations with mixed confidence.

The expected return for these experiments is shown in Table~\ref{tab:mujoco}. \textbf{Ours} outperforms \textbf{GAIL} and \textbf{DCC}, and is comparable with the \textbf{Oracle} in both environments. Also, \textbf{Ours} outperforms \textbf{Ours-Confidence}, which demonstrates the efficacy of multi-step partial trajectory matching. \textbf{Ours-Confidence} outperforms \textbf{Ours-Feature} and \textbf{Ours-Feature} outperforms \textbf{GAIL}, which demonstrate the efficacy of confidence-level and feature-level matching respectively. For 2-joint reacher and 5-leg ant, the highest p-values comparing with baselines (between \textbf{Ours} and \textbf{DCC}) are $0.147$ and $0.031$ (statistically significant) respectively.

\begin{table}[t]
\centering
    \renewcommand\tabcolsep{3pt}
    \resizebox{.3\textwidth}{!}{
    \begin{tabular}{c|c|c}
    \toprule
       & \multicolumn{1}{c|}{\textbf{Reacher}} & \multicolumn{1}{c}{\textbf{Ant}} \\
    \toprule
    \textbf{GAIL} &-47.13$\pm$5.67 &507.85$\pm$338.80\\
    \textbf{DCC} &-41.11$\pm$5.35 & 1218.59$\pm$231.74\\
    \textbf{Ours-Feature} &-43.97$\pm$4.68 &1059.02$\pm$280.95  \\
    \textbf{Ours-Confidence} &-41.82$\pm$6.80 &1263.04$\pm$343.02 \\
    \textbf{Ours}&\textbf{-39.53}$\pm$4.32 & \textbf{1556.43}$\pm$159.12\\
    \midrule
    \textbf{Oracle} &-28.73$\pm$3.97 & 1622.55$\pm$179.43\\
    \bottomrule
    \end{tabular}
    }
    \captionof{table}{\small{The expected return for MuJoCo environments. 
    }}
    \label{tab:mujoco}
\end{table}

\begin{figure}[t]
\vspace{-10pt}
    \centering
    \subfigure{\includegraphics[width=.25\textwidth]{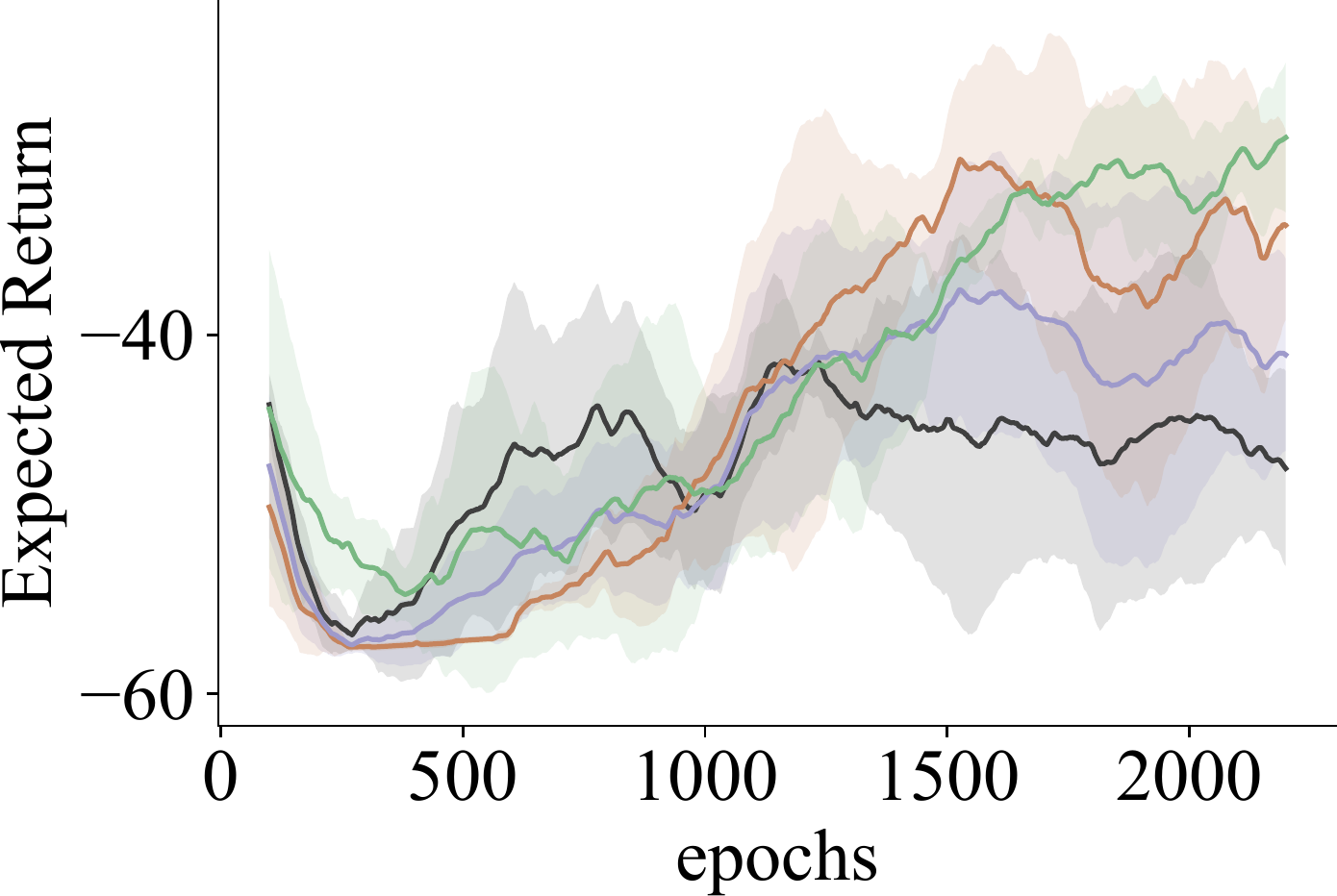}\label{fig:reacher_plot}}
    \hfill
    \subfigure{\includegraphics[width=.18\textwidth]{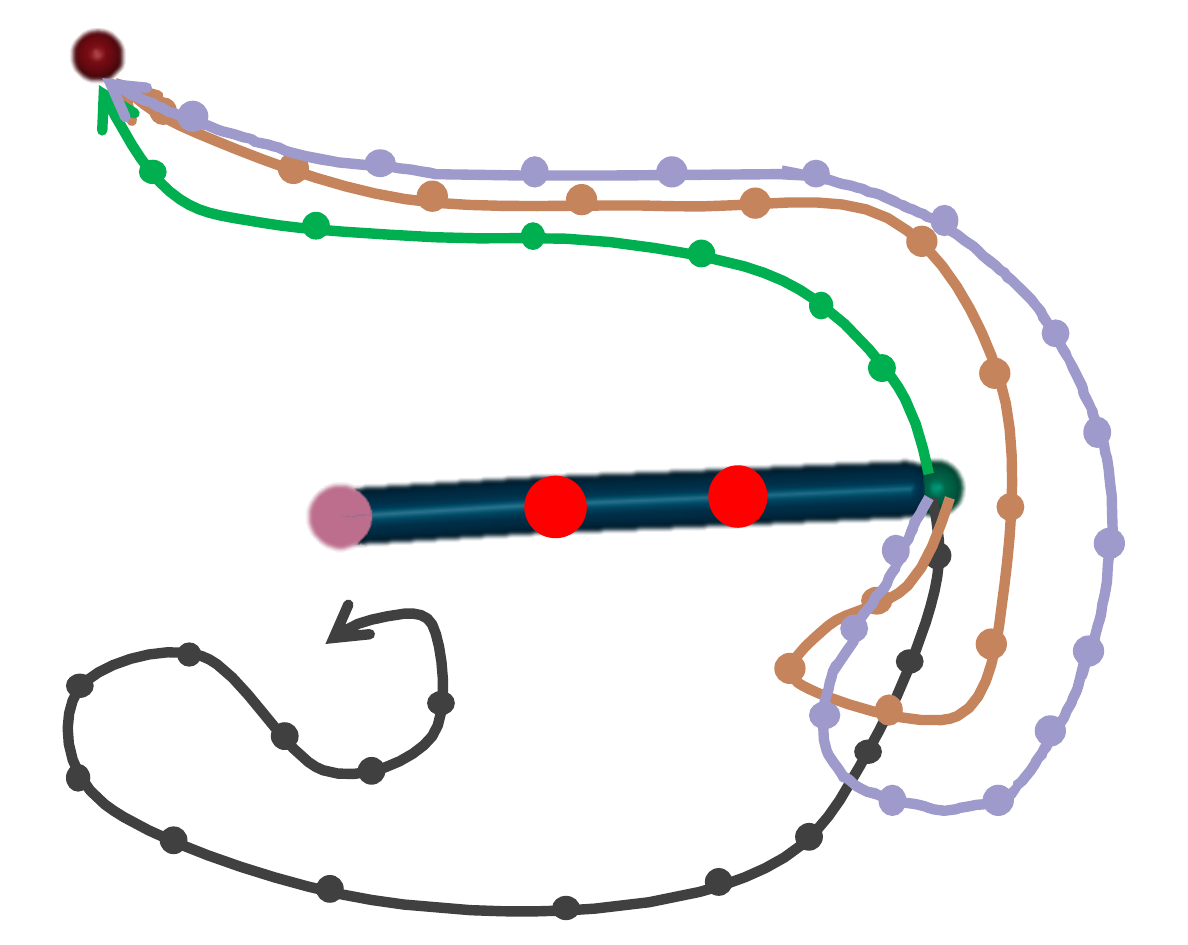}\label{fig:reacher_traj}} \\ 
    \subfigure{\includegraphics[width=.25\textwidth]{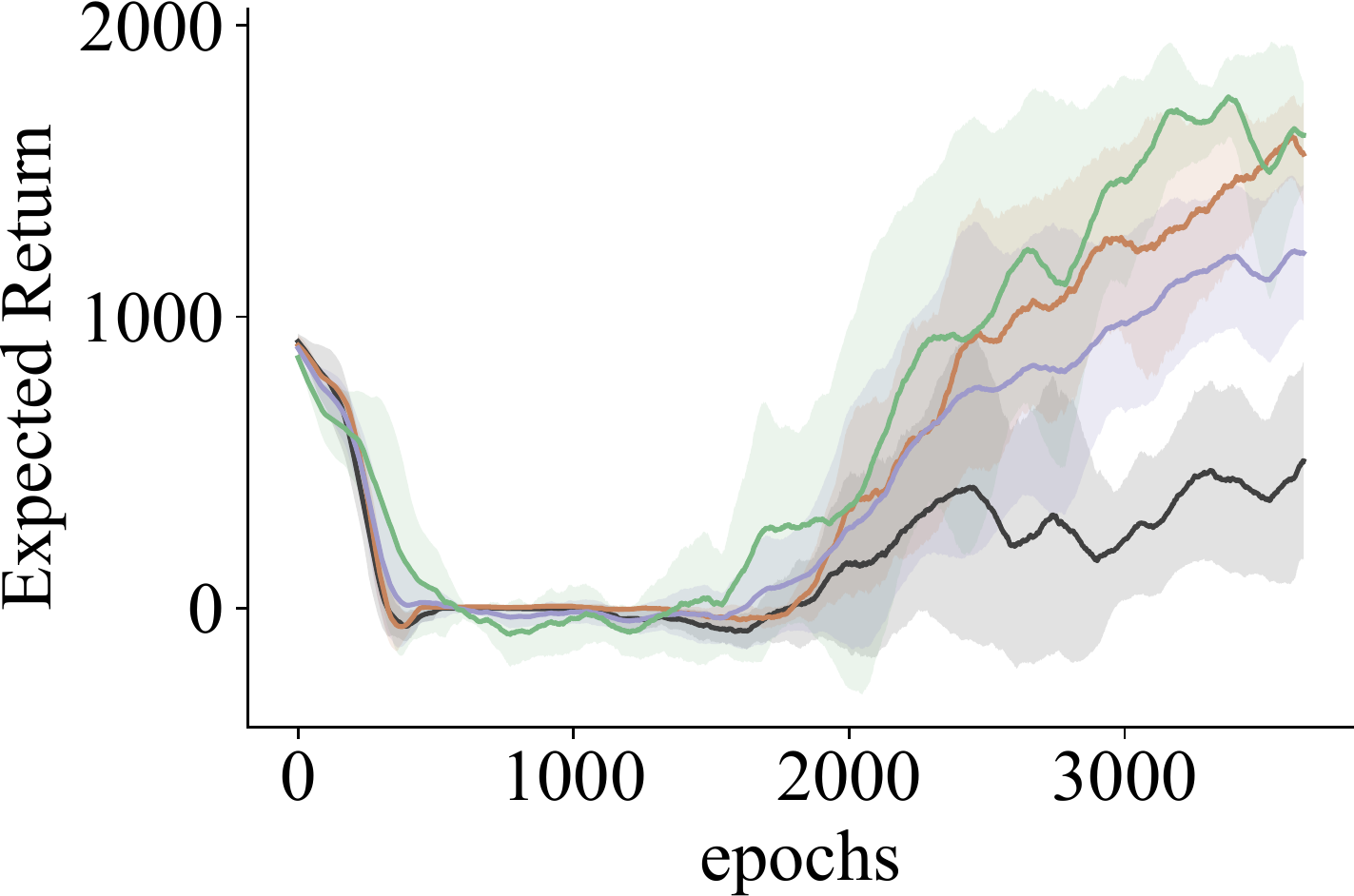}\label{fig:ant_plot}}
    \subfigure{\includegraphics[width=.23\textwidth]{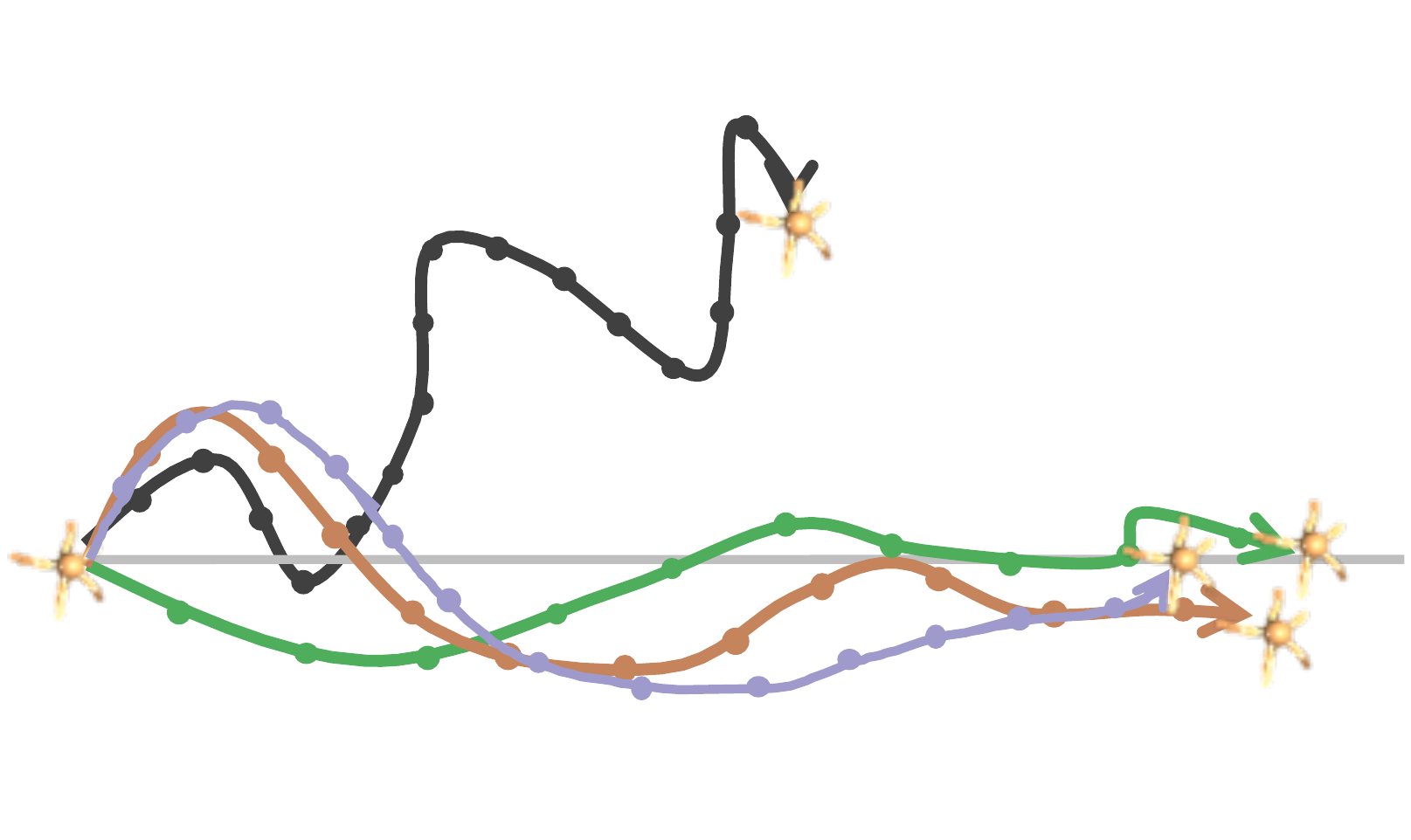}\label{fig:ant_traj}}
    \subfigure{\includegraphics[width=.49\textwidth]{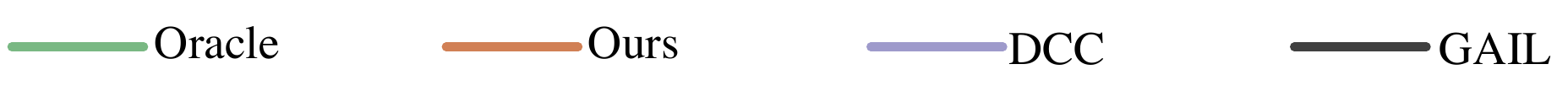}\label{fig:legend1}}
    \vspace{-20pt}
    \caption{\small{The top row and the bottom row are the expected return and sample trajectories for 2-joint reacher and 5-legged ant respectively. The light grey line in the Ant environment is the positive x-axis, which is the direction the ant is supposed to move toward.}}
\label{fig:mujoco_results}
    \vspace{-15pt}
\end{figure}
Figure~\ref{fig:reacher_plot} and~\ref{fig:ant_plot} show the learning curves of the algorithms. We observe that all methods have similar convergence rates.
Figure~\ref{fig:reacher_traj} and~\ref{fig:ant_traj} show a rollout generated by each of the policies. The rollout from the oracle policy reaches the goal in the most effective manner following the shortest trajectory length. \textbf{Ours} is the next closest to the \textbf{Oracle}, while the \textbf{GAIL} baseline fails at reaching the goal position at times.

\begin{minipage}{0.47\textwidth}
\vspace{5pt}
\begin{minipage}{.3\textwidth}
    \centering
    \includegraphics[width=\textwidth]{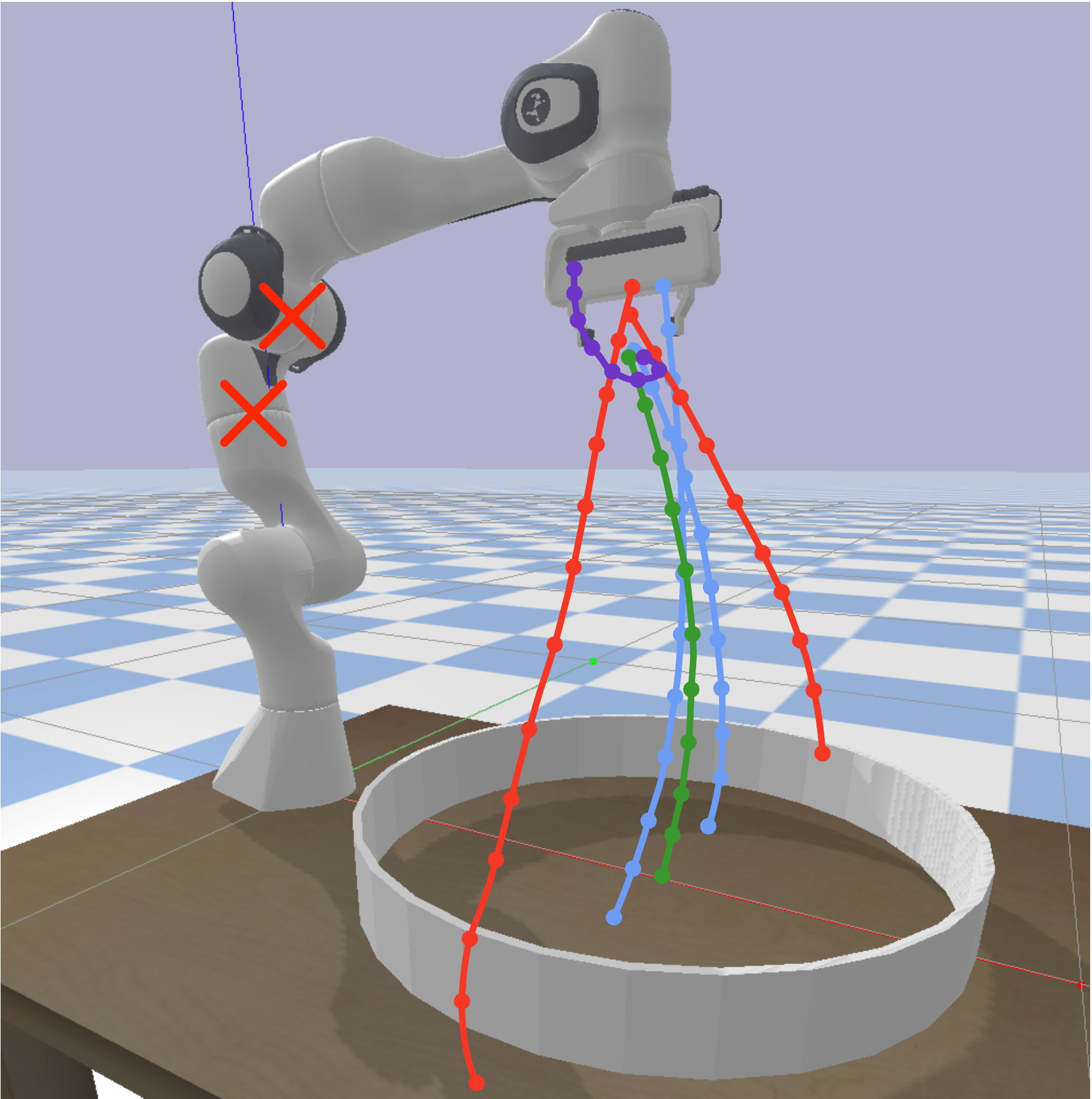}
    \captionof{figure}{\small{Illustration of different trajectories and the 5 DoF robot arm.}}
    \label{fig:simulated_robot}
\end{minipage}
\begin{minipage}{.7\textwidth}
    \centering
    
    \resizebox{\textwidth}{!}{
    \renewcommand\tabcolsep{2pt}
    \begin{tabular}{c|cc|cc}
    \toprule
    \multirow{2}{40pt}{Method}   & \multicolumn{2}{c|}{\textbf{OT}} & \multicolumn{2}{c}{\textbf{OSC}} \\
    \cmidrule{2-5}
    & Return & Success & Return & Success \\
    \midrule
    \textbf{GAIL} & -902.2$\pm$152.2 & 42.8$\pm$11.9 & -1065.8$\pm$105.8  & 21.2$\pm$8.6\\
    \textbf{DCC} & -495.6$\pm$46.5 & 76.4$\pm$3.4 & -723.7$\pm$121.4 & 49.5$\pm$15.2 \\
    \textbf{Ours-Feature} & -1198.4$\pm$80.4 & 31.2$\pm$7.5  & -868.1$\pm$265.7 & 37.6$\pm$18.3 \\
    \textbf{Ours-Confidence} & -781.9$\pm$12.7 & 59.6$\pm$0.8 & -821.2$\pm$121.4  & 42.0$\pm$11.9 \\
    \textbf{Ours} & \textbf{-154.8}$\pm$12.8 & \textbf{92.8}$\pm$1.0 & \textbf{-702.2}$\pm$104.4 & \textbf{52.4}$\pm$7.9 \\
    \midrule
    \textbf{Oracle} & -55.6$\pm$2.3 & 100.0$\pm$0.0 & -613.9$\pm$159.4 & 59.6$\pm$13.2 \\
    \bottomrule
    \end{tabular}
    }
            \captionof{table}{\small{The expected return and the success rate among $500$ trials (\%) of GAIL, DCC, Ours-Single, Ours w/o $\mathcal{L}_\text{con}$, Ours, and Oracle for the two setups OR and OSC in the simulated robot experiments.}}
            \label{tab:simulated_robot}
\end{minipage}
\vspace{5pt}
\end{minipage}

\noindent \textbf{Simulated Robot.}
As shown in Fig.~\ref{fig:simulated_robot}, we create a task to move the end-effector of a 5 DoF Panda Franka Robot arm toward the center of a plate without colliding with the boundaries. 
The reward function consists of the negative L2 distance to the center; collisions with the wall or failing to reach the table within the time limit; and positive reward when the end-effector reaches the plate. 

The source robot arm has 7 DoF and the target has 5 DoF, where the joints marked by red in Fig.~\ref{fig:simulated_robot} are disabled. We show different kinds of trajectories: (green--optimal trajectory reaching the center of the plate), (blue--suboptimal trajectories reaching the plate but not the center), (red--failure cases colliding with the wall), (violet--another type of failure case failing to reach the plate within the time limit). We create two settings: (1) (OT) consists of the optimal (green) and out-of-time (violet) trajectories; and (2) (OSC) consists of optimal (green), suboptimal (blue), and collision (red) trajectories. In the OT setting, we use the joint and end-effector position as the state, which serves as an easy setup since the end-effector position is shared between the 7-DoF and the 5-DoF. In the OSC setting, we use the joint position, velocity and torque as the state, which is difficult to align. For both settings, we use the 3D end-effector position difference as the action space.

Expected return and the success rate are shown in Table~\ref{tab:simulated_robot}. \textbf{Ours} outperforms \textbf{GAIL} and \textbf{DCC} with a large margin, which indicates that the learned target confidence predictor can assign higher confidence to useful demonstrations than \textbf{DCC}, and \textbf{GAIL}. The order of the performance from high to low is \textbf{Ours}, \textbf{Ours-Confidence}, \textbf{Ours-Feature}, and \textbf{GAIL}, which demonstrates the efficacy of our multi-length partial trajectory matching, confidence-level matching and feature-level matching respectively. The highest p-values when comparing with baselines (between \textbf{Ours} and \textbf{DCC}) are $5.64\times10^{-12}$ for the success rate and $1.25\times 10^{-13}$ for the expected return in the OR setting, and $0.006$ for the success rate and $0.031$ for the expected return in the OSC setting, demonstrating statistical significance.

\begin{figure}[h]
    \vspace{-10pt}
    \centering
    \subfigure[Simulated Robot ]{\includegraphics[width=.18\textwidth]{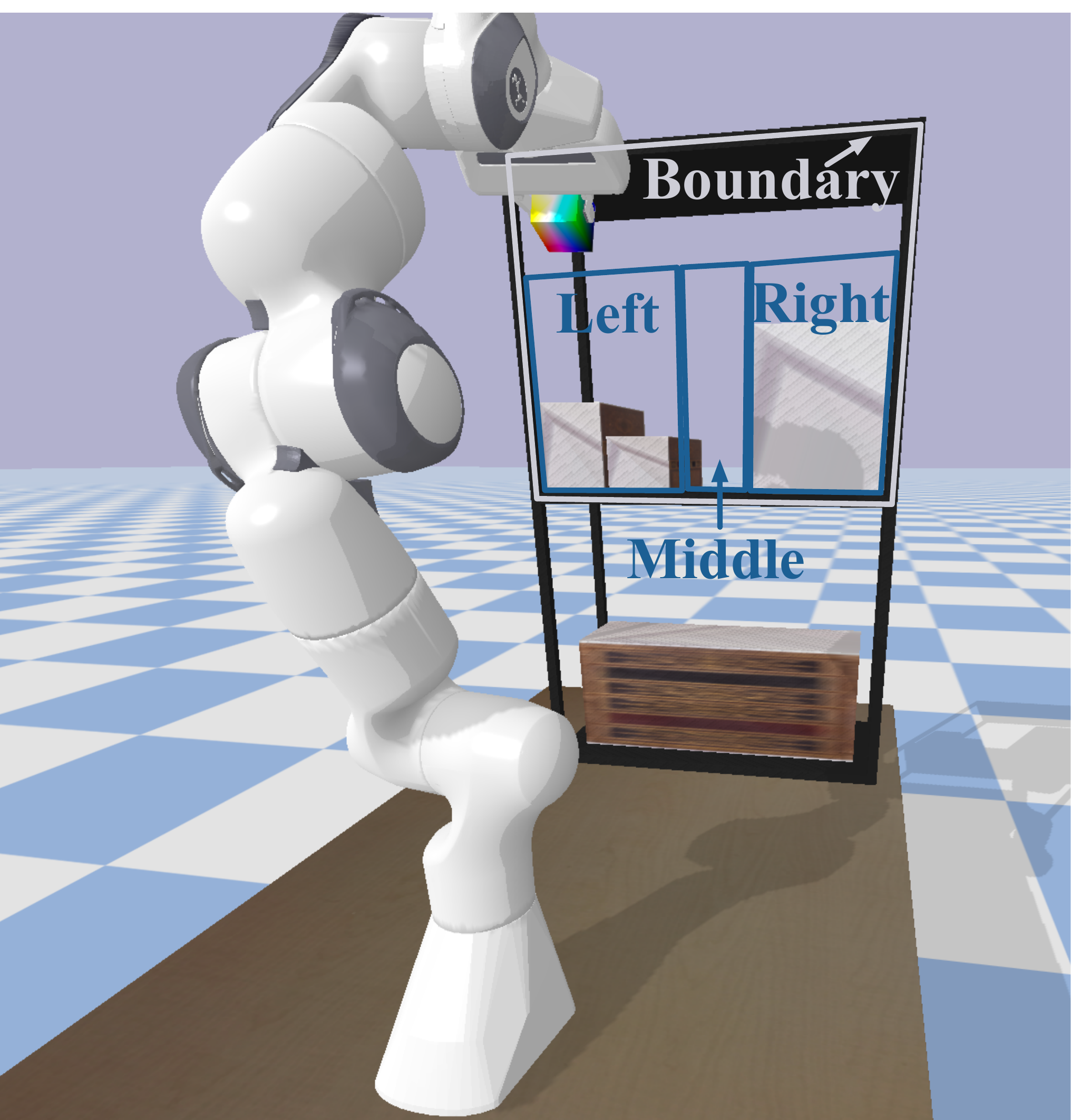}\label{fig:sim_to_real_sim}}
    \subfigure[Real Robot ]{\includegraphics[width=.2\textwidth]{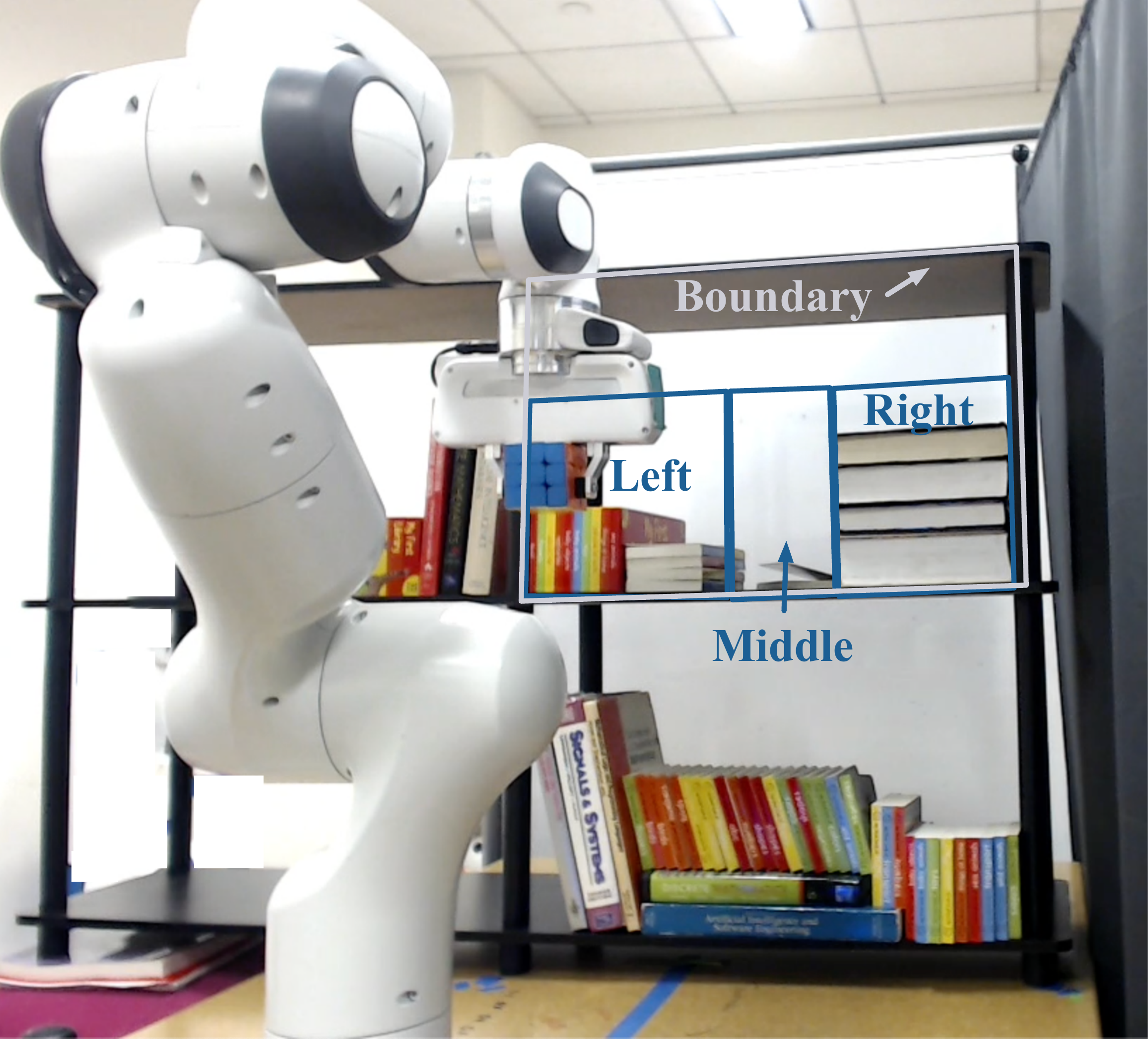}\label{fig:sim_to_real_real}} \\
    \subfigure[Expected Return]{\includegraphics[width=.2\textwidth]{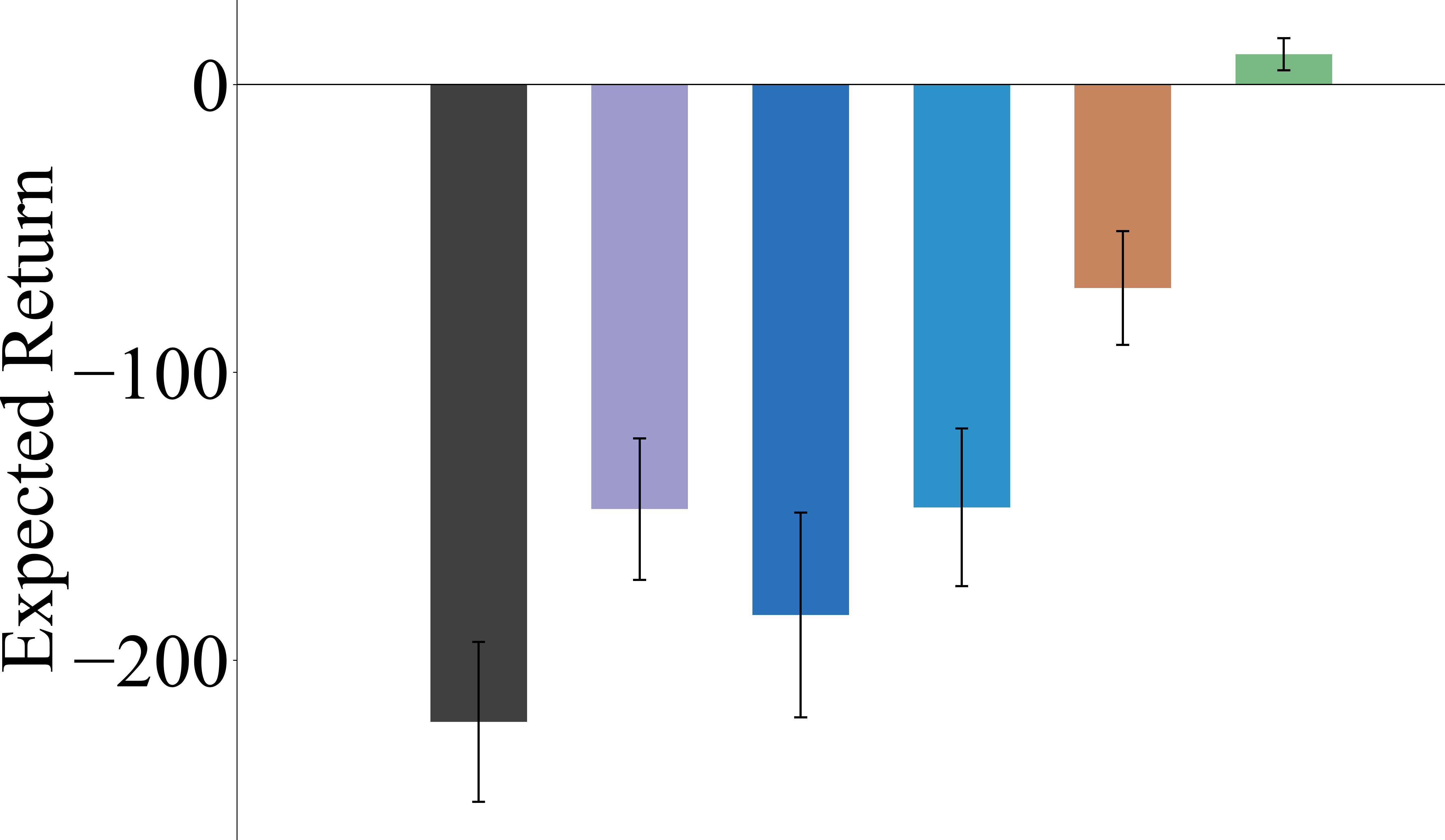}\label{fig:sim_to_real_result_return}}
    \subfigure[Success Rate]{\includegraphics[width=.2\textwidth]{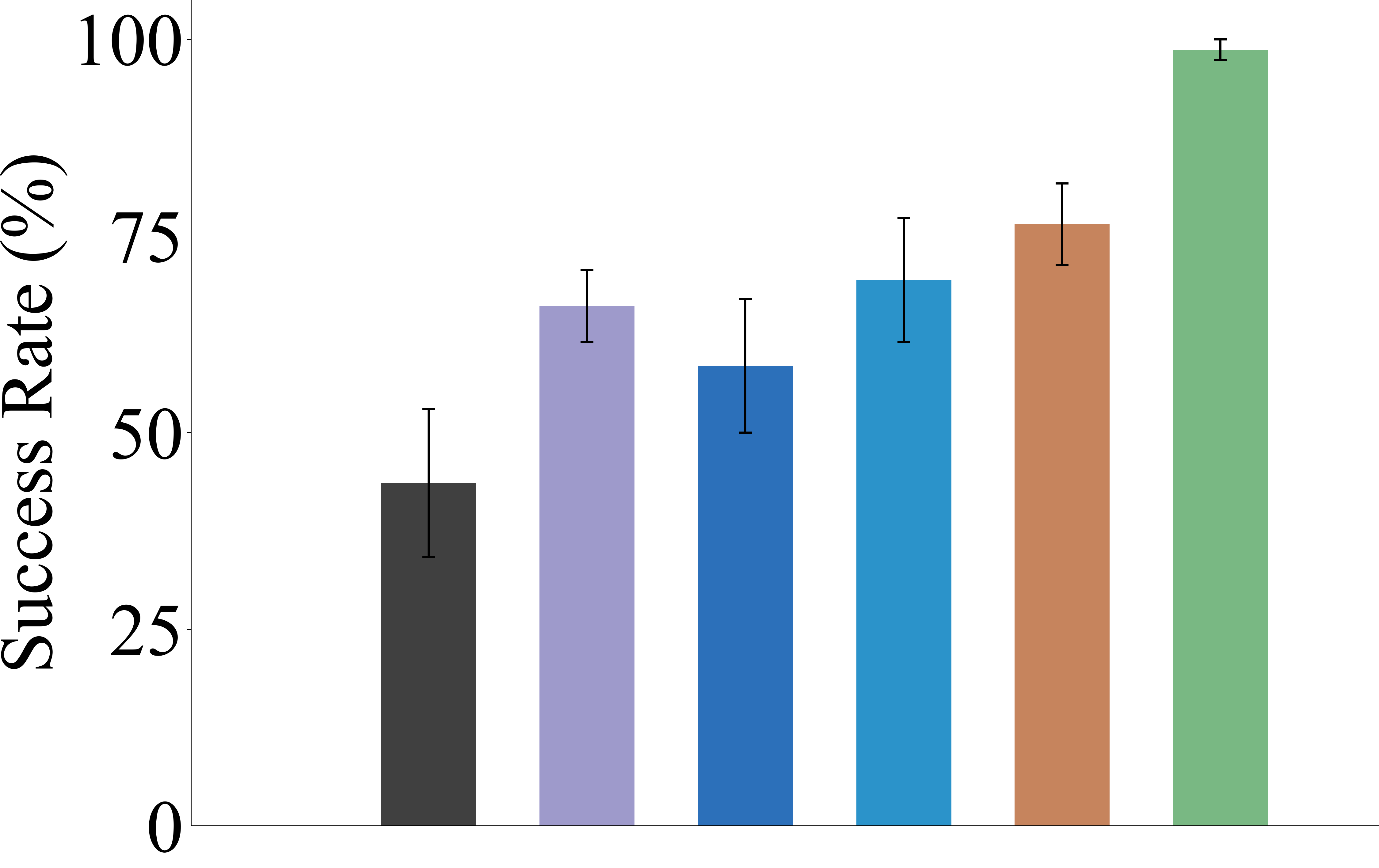}\label{fig:sim_to_real_result_rate}}
    \subfigure{\includegraphics[width=.49\textwidth]{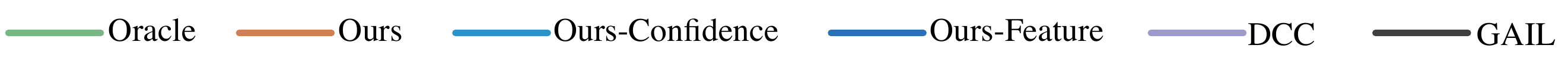}\label{fig:legend}}
    \vspace{-20pt}
    \caption{\small{(a-b) Illustration of the simulated robot and the real robot arm environments. Different colors indicate different areas to place the objects. (c-d) Expected return and success rate.}}
    \label{fig:sim_to_real}
    \vspace{-5pt}
\end{figure}

\noindent \textbf{Sim-to-Real Environment.}
In the sim-to-real environment, we use a simulated and a real Franka Panda Arm as the source and the target respectively (both with 7-DoF). The task is to move the cube to the upper layer of the shelf (see Fig.~\ref{fig:sim_to_real}). On the shelf, there is a large stack of books on the right and a small stack on the left. The middle area is empty. We assign positive rewards to success of placing the cube on the shelf, with rewards $200$, $300$, and $100$ for placing on the left, the middle and the right respectively. We penalize time by giving a $-1$ reward for each time step. The average number of steps for all demonstrations is about $300$. If the arm fails to put the cube within $1000$ steps, it receives reward $0$. We use the joint position, velocity and the end effector position as the state space and joint forces as the action space.

Fig.~\ref{fig:sim_to_real_result_return} and~\ref{fig:sim_to_real_result_rate} show \textbf{Ours} outperforms baselines \textbf{DCC} and \textbf{GAIL}. The performance from high to low is: \textbf{Ours}, \textbf{Ours-Confidence}, \textbf{Ours-Feature}, and \textbf{GAIL}, which demonstrates that multi-length partial trajectory matching, feature-level and confidence-level matching are all necessary to learn a common latent space and an accurate target confidence predictor for the real robot. The p-value between \textbf{Ours} and the highest baseline, \textbf{DCC}, is $0.0004$ for the expected return and $0.0052$ for the success rate, demonstrating statistical significance.  
\vspace{5pt}

\noindent \textbf{Varying Composition of Demonstrations.}
We conduct experiments by varying the composition of source demonstrations but fixing the target demonstrations in the two Mujoco environments to test the robustness of confidence predictor under different demonstration mixtures. We do not change the target demonstrations since that will also influence the imitation learning performance and we cannot test the efficacy of the confidence predictor. In the \emph{Demonstration} column in Table~\ref{tab:reacher_ant_varying}, we show the number of demonstrations that are collected from different policies from random to optimal in Reacher (3 policies) and Ant (5 policies). 
From top to bottom, the optimality of source and target demonstrations deviates more and more. We observe that the performance drops with larger deviation but does not drop too much even on the last row. This demonstrates that the proposed approach can work stably even with confidence distribution shift between demonstrations. 


\begin{minipage}{.47\textwidth}
\vspace{5pt}
    \centering
    
    \resizebox{\textwidth}{!}{
    \renewcommand\tabcolsep{2pt}
    \begin{tabular}{c|c||c|c}
    \toprule
    \multicolumn{2}{c||}{\textbf{Reacher}} & \multicolumn{2}{c}{\textbf{Ant}} \\

    Demonstration & Expected Return & Demonstration & Expected Return \\
    \midrule
     1-5-94 & \textbf{-36.69}$\pm$5.65 & 48-49-97-5-1  & \textbf{1525.03}$\pm$176.91 \\
     1-49-50 & -37.38$\pm$10.11 & 48-73-73-5-1 & 1489.85$\pm$268.22 \\
     1-94-5 & -37.96$\pm$6.61  & 48-97-49-5-1 & 1436.78$\pm$176.91 \\
    47-48-5 & -37.61$\pm$5.50 & 72-73-49-5-1  & 1351.44$\pm$115.59\\
     94-1-5 & -39.49$\pm$2.56 & 97-48-49-5-1 & 1345.96$\pm$299.08 \\
    \midrule
    \bottomrule
    \end{tabular}
    }
    \vspace{-5pt}
            \captionof{table}{\small{Expected return with respect to varying compositions of the source demonstrations for Reacher and Ant. The numbers in the demonstration column indicate the number of the demonstrations collected from random to optimal policies (3 policies for Reacher and 5 policies for Ant).}}
            \label{tab:reacher_ant_varying}
\end{minipage}


\section{Conclusion}
\label{sec:conclusion}
\noindent \textbf{Summary.} We propose an algorithm to learn from imperfect demonstrations, where the demonstrations can be suboptimal or even fail. 
We learn a confidence predictor by leveraging confidence labels and demonstrations in a different but correspondent environment.
We show the policy learned by our method outperforms other baselines in various environments.

\noindent \textbf{Limitations and Future Work.} 
We require the source and target environments to be correspondent.
In the future, we plan to relax this assumption by defining weaker correspondence only between confidence instead of the strict definition of correspondence on dynamics.
\section{Acknowledgements}
We would like to thank FLI grant RFP2-000, NSF Award Number 1849952 and 1941722, and ONR for their support.
\clearpage





{\small
\bibliography{example.bib}
\bibliographystyle{IEEEtran}
}

\end{document}


\maketitle
\thispagestyle{empty}
\pagestyle{empty}

\appendix
In the Appendix, we provide the details on the algorithm and further explain experiment details in Sec.~\ref{sec:exp_detail}. 
The videos are shown on our \href{https://sites.google.com/view/adversarialconfidencetransfer
}{\color{RedOrange}{website}}.

\section{Algorithm}\label{sec:algo}
We go through the steps of the algorithm of learning from imperfect demonstrations via adversarial confidence transfer in Algorithm~\ref{alg:algo}. 
Lines 2-6 show the first stage of our framework as in Fig. 2 (left): training $E^\text{src}$ with source confidence-labeled data. Lines 7-14 show the process of the second stage in Fig. 2 (left): aligning the distribution of source and target state-action pairs in the common latent space. Lines 8-11 and Line 12-14 show we iteratively update the target encoder $E^\text{tar}$ and the discriminators $D_k$ and $D'_k$.

After we learn the confidence prediction function $F(E^\text{src})$, we predict the confidence for each state-action pair in the target demonstrations and conduct standard imitation learning on the re-weighted target demonstrations similar to the approach described in [6]. Specifically, we use Generative Adversarial Imitation Learning as our imitation learning algorithm. The loss can be derived as
\begin{equation}
    \begin{aligned}
    \min_\theta \max_\omega &\mathbb{E}_{(s,a)\sim \pi_\theta}[ D_\omega(s,a)] \\
 &+ \mathbb{E}_{(s,a)\sim \Xi^\text{tar}}[F(E^\text{src}(s,a)) (1-D_\omega(s,a))],
    \end{aligned}
\end{equation}
where the $\pi_\theta$ is the policy parameterized by $\theta$ and $D_\omega$ is the discriminator parameterized by $\omega$.

\begin{algorithm*}[htbp]
\KwIn{The demonstration set of the source environment $\Xi^\text{src}$ and of the target environment $\Xi^\text{tar}$. The confidence function $c^\text{src}$ for the source environment.}
  Initialize $E^\text{src}$, $E^\text{tar}$, $F$, $D_i-D_K$ and $D'_i-D'_k$\;
  \While {not converging}{
 Sample a batch of state-action pairs $\{(s^\text{src}, a^\text{src})\}$ from $\Xi^\text{src}$\;
 Compute the confidence for state-action pairs in $\{(s^\text{src}, a^\text{src})\}$ with $c^\text{src}$\;
 Train $E^\text{src}$ and $F$ with $\{((s^\text{src}, a^\text{src}), c^\text{src}(s^\text{src}, a^\text{src}))\}$ according to the loss in Eqn. (1) and the optimization objective in Eqn. (4)\;
 }
  Fix the parameters of $E^\text{src}$ and $F$. 
  \While {not converging}{
  \For{$k=1\rightarrow K$}{
  Sample a batch of partial trajectories $\{(s_1,a_1,\dots,s_k,a_k)\}$ with length $k$\;
  }
  Train $E^\text{tar}$ with partial trajectories of all lengths: $\{(s_1,a_1,\dots,s_k,a_k)\}|_{k=1,\cdots,K}$ according to the loss in Eqn. (2) and (3), and the optimization objective in Eqn. (6)\;
  \For{$k=1\rightarrow K$}{
  Train $D_k$ and $D'_k$ with the partial trajectories $\{(s_1,a_1,\dots,s_k,a_k)\}$ according to the loss in Eqn. (2) and (3), and the optimization objective in Eqn. (5)\;
  }
  }
  \KwOut{The target confidence predictor $F\circ E^\text{tar}$.}
  \caption{Algorithm for learning the target confidence predictor}\label{alg:algo}
\end{algorithm*}

\section{Experimental Details}\label{sec:exp_detail}

\subsection{Implementation Details}
Our approach uses the source and target demonstrations and the source confidence function as the input. To implement the baseline methods we make the following choices: For GAIL~\cite{ho2016generative}, we directly conduct imitation learning from the demonstrations in the target environment without any confidence weights. For DCC~\cite{zhang2021learning}, the original paper uses trajectories sampled from a random policy to learn the translation mappings. In our setting, we use both the demonstrations and the random trajectories to learn the translation mapping. 

To generate the ground-truth confidence score for the source demonstrations, following the common practice in prior works~\cite{pmlr-v97-wu19a,cao2021learning}, we normalize the expected return of all the demonstrations to $[0,1]$ by min-max normalization and assign all the state-action pairs in each trajectory with the confidence of the trajectory.

\subsection{Mujoco}\label{sec:mujoco}
\noindent \textbf{Demonstration Generation.} For the Reacher environment, we compute the median value between the expected returns of a random policy and the optimal policy and find a partially-trained policy to match the value. Then we have three policies: random policy, 50\% partially-trained policy, and optimal policy. We generate $94$, $5$ and $1$ trajectories from the random policy, the 50\% partially-trained policy, and the optimal policy respectively, which are $100$ trajectories in total.

For the Ant environment, we select five values with an equal interval between the expected returns of a random policy and the optimal policy and find three partially-trained policies to match the three values. Then, we have five policies: random policy, 25\% partially-trained policy, 50\% partially-trained policy, 75\% partially-trained policy and optimal policy. We generate $48$, $49$, $97$, $5$ and $1$ trajectories from the random policy, the 25\% partially-trained policy, the 50\% partially-trained policy, the 75\% partially-trained policy, and the optimal policy respectively.

\subsection{Simulated Robot Arm} 
\noindent \textbf{Reward Design.} We use $x_\text{center}$ to denote the center of the circle on the table surface, and $x_\text{t}$ to denote the position of the end-effector at time step $t$. Based on this goal, we develop a reward function consisting of three parts: the first part is the negative L2 distance between the $x_\text{t}$ and $x_\text{center}$; the second part penalizes collisions with $-1000$ reward when the end-effector collides with the wall or does not reach the table within the time limit $T_h$; and the third part is a gain of positive reward when the end-effector reaches the table surface within the circle area and the closer to the center, the higher the reward. The reward for the simulated robot arm environment is defined as $R_t=-\lVert x_t-x_\text{center} \rVert_2-1000\times \left(\mathbb{I}\left[\text{collide}\right] |\mathbb{I}\left[t\ge T_h\right]\right)+400\times\exp(-\frac{\lVert x_\text{reach}-x_\text{center} \rVert_2}{0.1})\times\mathbb{I}\left[\text{reach}\right]$. 
The initial position of the joint that controls the rotation of the hand can vary freely in $[0,2\pi]$, and the initial position of all the other joints of the arm can vary $\pm 0.3$. We create an initial position area within these regions to allow some variance for the task while not exceeding the joint limits of the robot.

\noindent \textbf{Demonstration Generation.} We create out-of-time trajectories by injecting noisy actions with some probability and we vary the probability to create different kinds of out-of-time trajectories. We vary the suboptimal trajectories by changing the distance from the goal point on the table to the circle center.

\noindent \textbf{Composition of Demonstrations.} In both the source and target environments, we use the same composition of trajectories. In the OR setting, we collect $10$ optimal trajectories and $200$ out-of-time trajectories. In the OSC setting, we collect $40$ optimal trajectories, $80$ suboptimal trajectories, and $150$ collision failure trajectories.
The composition is designed to contain more suboptimal demonstrations and failure cases to make the imitation learning problem more challenging.

\noindent \textbf{Implementation Details.} To implement our approach in the simulated robot arm and the real robot arm, we use a three-layer fully-connected network for $E^\text{src}$ and $E^\text{tar}$ respectively, a one-layer fully-connected network for $F$ and a three-layer fully-connected network for $D_k$ and $D'_k$ ($k=1,\cdots,K$) respectively. We use behavior cloning~\cite{bain1995framework} as the imitation learning algorithm to learn the final policy from the reweighted demonstrations. 

\subsection{Sim-to-Real Environment}
In the Sim-to-Real environment, the network and the imitation learning algorithm are the same as the simulated robot arm environment.

\noindent \textbf{Composition of Demonstrations.} For the simulated robot arm, we collect human demonstrations by controlling the end-effector by the mouse. For the real robot arm, we collect human demonstrations by moving the robot arm by hand. For both the simulated and real robot arms, the demonstrations are composed of $5$ trajectories for placing the cube on the left, middle and right each, $5$ trajectories colliding with the left, up, bottom boundary of the upper shelf each, and $5$ trajectories colliding with the books on the right side (we cannot reach the right boundary due to the joint limit).

{\small
\bibliography{example.bib}
\bibliographystyle{IEEEtran}
}